\documentclass{article}
\usepackage{arxiv}
\usepackage[utf8]{inputenc} 
\usepackage[T1]{fontenc}    
\usepackage{cite}
\usepackage{amsmath,amssymb,amsfonts}
\usepackage{algorithm}
\usepackage{algorithmic}
\usepackage{graphicx}
\usepackage{textcomp}
\usepackage{xcolor}
\usepackage{multirow}
\usepackage{colortbl}
\def\BibTeX{{\rm B\kern-.05em{\sc i\kern-.025em b}\kern-.08em
    T\kern-.1667em\lower.7ex\hbox{E}\kern-.125emX}}
\begin{document}

\title{Robust and Efficient Noisy-Label Time-Series Classification via Dynamic Time Warping Based Granular Ball Computing\\
}
\author{
Ziqiang Li$^{1,3,*}$ \\
\texttt{ziqiang-li@nitech.ac.jp}
\And
Yun Liu$^{2}$ \\
\texttt{liuyun@kit.ac.jp}
\And
Gouhei Tanaka$^{1,3}$ \\
\texttt{tanaka.gouhei@nitech.ac.jp}
\\[1ex]
$^{1}$ Department of Computer Science, Graduate School of Engineering, \\
Nagoya Institute of Technology, Nagoya, 466-8555, Japan \\
$^{2}$ Faculty of Information and Human Sciences, \\
Kyoto Institute of Technology, Kyoto, 606-8585, Japan\\
$^{3}$ International Research Center for Neurointelligence, \\
The University of Tokyo, Tokyo, 113-0033, Japan.
}
\maketitle
\begin{abstract}
Dynamic Time Warping (DTW)-based Nearest-Neighbor (NN) classifiers are effective for time-series classification but are vulnerable to mislabeled training samples and require numerous DTW computations during inference. We propose DTW-based Granular Ball Computing (DTW-GBC), which organizes temporally similar training samples into granular balls and performs classification at the granule level. We further develop two granular-ball construction strategies for DTW-GBC. Experiments on four benchmark datasets with symmetric label noise show that the two DTW-GBC variants generally mitigate the performance degradation caused by label noise while requiring substantially fewer comparisons than DTW-based 1-NN during inference. These findings suggest that DTW-GBC provides a favorable balance between classification robustness and inference efficiency.
\end{abstract}

\section{Introduction}
Time-series classification (TSC) is a fundamental task in time-series analysis and has been widely applied in various real-world domains~\cite{faouzi2022time}. Most supervised TSC methods rely on accurately labeled training data to identify discriminative temporal patterns. In practical applications, however, obtaining reliable labels is often difficult and expensive. Label noise may arise from subjective or inconsistent human annotations, insufficient domain expertise, ambiguous temporal patterns, or imperfect automatic labeling systems~\cite{frenay2014classification}. Such mislabeled samples can mislead classification models and substantially degrade their inference performance~\cite{ma2023ctw,liu2023scale}. Therefore, developing TSC methods that are robust to label noise remains an important research problem.
\begin{figure*}[htbp]
    \centering
    \includegraphics[scale=0.8]{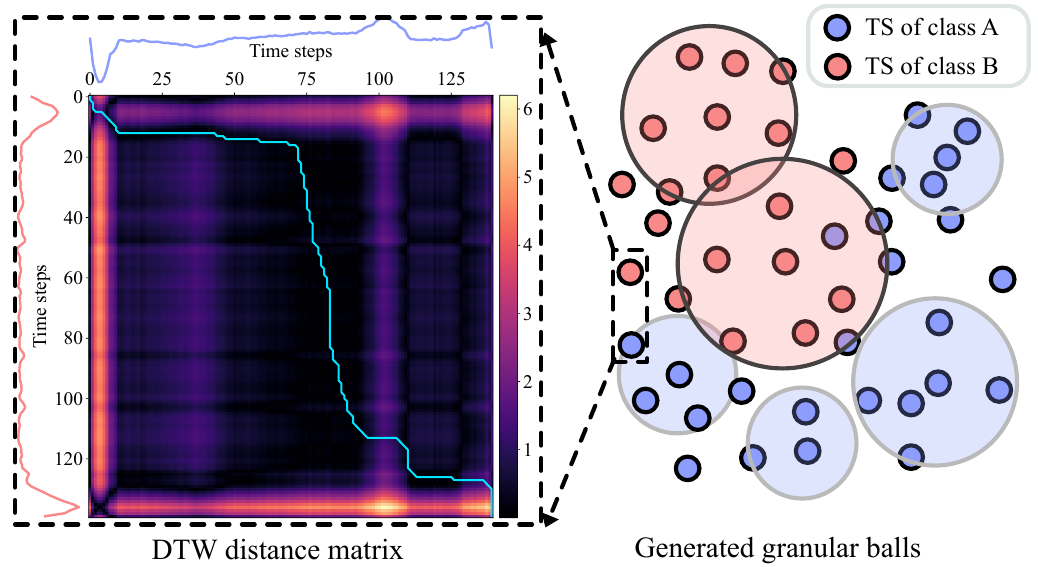}
    \caption{Overview of the proposed DTW-GBC framework. DTW is first used to compute pairwise distances between training time series, with the DTW distance matrix and optimal warping path for one sample pair illustrated on the left. These pairwise distances are then used to recursively organize the training samples into granular balls, as illustrated on the right, for subsequent granular-ball-level nearest-neighbor classification.}
    \label{fig:DTW-GBC-framework}
\end{figure*}
\par
Among existing TSC approaches, distance-based classifiers provide a simple yet effective solution by assigning labels according to the similarity between test and training samples~\cite{abanda2019review}. In particular, Dynamic Time Warping (DTW)-based $1$-nearest-neighbor ($1$-NN) classification is widely adopted because DTW can capture similar temporal patterns despite local phase shifts, speed variations, and unequal sequence lengths~\cite{sakoe1978dynamic}. Nevertheless, this instance-level classification mechanism presents two major limitations. First, because $1$-NN directly inherits the label of the nearest training sample, a mislabeled nearest neighbor can immediately result in an incorrect classification. Second, each test sample must be compared with all training samples, and each DTW comparison requires dynamic programming over the two sequences. Consequently, DTW-based $1$-NN may suffer from both degraded robustness under label noise and high computational cost during inference.
\par
Granular Ball Computing (GBC) has recently emerged as an adaptive multi-granularity representation paradigm that organizes individual samples into a compact set of granular balls~\cite{xia2023granular}. This granule-level representation is potentially suitable for addressing the two limitations of DTW-based $1$-NN. Assigning labels at the granule level can reduce the direct influence of isolated mislabeled samples, while replacing individual training samples with a smaller number of granular balls can reduce the number of comparisons required during inference. However, existing GBC methods are generally designed for static data, where granular balls are constructed using geometric operations in a fixed-dimensional feature space~\cite{xia2019granular}. Directly extending these methods to time-series data is nontrivial because temporal misalignment, variable sequence lengths, and multivariate temporal structures must be considered when measuring sample similarity. Despite the effectiveness of DTW in measuring time-series similarity, the construction and recursive refinement of granular balls directly in the DTW distance space remain underexplored. Moreover, it remains unclear whether such a representation can simultaneously improve robustness to noisy labels and reduce the inference cost of DTW-based nearest-neighbor classification.
\par
To address these issues, we propose Dynamic Time Warping-Based Granular Ball Computing (DTW-GBC) for robust and efficient noisy-label time-series classification. DTW-GBC first computes pairwise DTW distances among the training samples and then recursively partitions them into granular balls in a coarse-to-fine manner. During inference, DTW-GBC performs 1-NN classification at the granular-ball level by comparing each test sample only with the generated granular balls, rather than with all individual training samples.
\par
The main contributions of this work are summarized as follows:
\begin{itemize}
    \item We propose DTW-GBC, which constructs granular balls directly in the DTW distance space and enables granule-level representation and classification of univariate and multivariate time series.
    \item We develop two medoid-based recursive splitting strategies, namely random splitting and label-informed splitting.
    \item Experiments on four benchmark datasets with symmetric label noise demonstrate that DTW-GBC generally alleviates the classification-performance degradation caused by noisy labels while requiring fewer comparisons during inference than the DTW-based 1-NN classifier.
\end{itemize}
\par
The rest of this paper is organized as follows: The related works  are introduced in Section~\ref{sec: related works}. Our proposal is described in Section~\ref{Sec: DTW-GBC}. The analysis of computational complexity is given in Section~\ref{sec: complexity analysis}. The details of experiments are presented in Section~\ref{Sec: Experiments}. The conclusion is given in Section~\ref{Sec: Discussion}.
\section{Related Works}
\label{sec: related works}
\subsection{Distance-based Time Series Classification}
Distance-based TSC methods constitute an important branch of TSC, where the distance between time series samples is measured using a predefined distance function, and class labels are typically assigned by nearest-neighbor-based classifiers~\cite{abanda2019review}. Among various distance measures, DTW is one of the most widely used elastic distance measures for time series data \cite{sakoe1978dynamic}. By allowing temporal alignment between two sequences, DTW can capture shape-based similarities even in the presence of local phase shifts or speed variations. Owing to this property, the combination of DTW and nearest-neighbor classifiers has been widely adopted as a strong baseline for TSC \cite{bagnall2017great}. However, since nearest-neighbor predictions directly depend on the labels of retrieved training samples, mislabeled neighbors may propagate incorrect class information and degrade classification performance. Moreover, DTW-based nearest-neighbor methods operate at the individual-sample level and require repeated pairwise distance computations, which can become computationally expensive as the number of time series increases \cite{rakthanmanon2012searching}. Therefore, improving both the robustness and efficiency of distance-based TSC while preserving the temporal alignment capability of DTW remains an important research issue.
\subsection{Granular Ball Computing}
GBC is an emerging granular computing paradigm that represents observed data using a set of Granular Balls (GBs) rather than individual samples~\cite{xia2023granular}. By recursively partitioning data in a coarse-to-fine manner, GBC can describe the original sample space with fewer compact data granules, thereby reducing computational cost while preserving the main data structure. Moreover, this granule-level representation can improve robustness to noise and outliers and provide interpretable local structures for downstream learning \cite{xia2023granular}. Due to these advantages, GBC has been applied to various machine learning tasks, including classification \cite{xia2019granular}, clustering \cite{xie2023efficient} and regression \cite{rastogi2026robust}. Currently, granular-ball computing methods for time series processing generally rely on a time series encoder (e.g. Echo State Networks~\cite{li2022multi}) followed by a pooling operation to transform temporal features into static feature representations~\cite{wang2026efficient}. Euclidean metric is then used to calculate distances and construct granular-ball representations~\cite{shen2026finding}. In contrast, methods that directly compute distances between raw time series for granular-ball construction remain underexplored.
\section{Dynamic Time Wraping based Granular Ball Computing}
\label{Sec: DTW-GBC}
In this section, we describe the details of the proposed DTW-GBC for time series classification. Figure~\ref{fig:DTW-GBC-framework} shows the overview of the proposed method. The key idea is to construct GBs based on distance between two different time series samples measured by DTW. 
\par For ease of the following description, we represent a multivariate time-series dataset by $\mathcal{S} = \left \{ \left ( \mathbf{X}^{(i)},y^{(i)} \right )  \right \} _{i=1}^{N}$, where the $i$-th element of $\mathcal{S}$ is a pair consisting of a matrix $\mathbf{X}^{\left ( i \right ) }  = \left [ \mathbf{x}^{\left ( i \right ) }_1, \mathbf{x}^{\left ( i \right ) }_2, \ldots, \mathbf{x}^{\left ( i \right ) }_{T^{(i)}} \right ] \in \mathbb{R}^{M\times T^{(i)}}$ and the corresponding label $y^{(i)}\in \mathbb{Z}$. The symbols $M$ and $T^{(i)}$ denote the dimensionality and sequence length of $\mathbf{X}^{(i)}$, respectively.
\subsection{Dynamic Time Warping}
To measure the difference between two multivariate time series $\mathbf{X}^{^{\left ( i \right ) }}$ and $\mathbf{X}^{^{\left ( j \right ) }}$, a distance matrix $\mathbf{D} \in \mathbb{R}^{T^{(i)} \times T^{(j)}}$ is built. The $(m,n)$-th element of $\mathbf{D}$, denoted by $\mathbf{D}_{\left [ m, n \right ]}$, represents the minimum cumulative distance required to align the first $m$ and $n$ elements of the two sequences, respectively. The $(m,n)$-th element of \(\mathbf{D}\) is computed recursively using dynamic programming, which can be formulated as follows:
\begin{equation}
\label{eq1}
\begin{aligned}
\mathbf{D}_{[m,n]}
={}& \left\lVert \mathbf{X}^{(i)}_{\left [ :,m \right ]}-\mathbf{X}^{(j)}_{\left [ :,n \right ]} \right\rVert_{2}^{2} \\
&+ \min\left(
\mathbf{D}_{[m-1,n]},
\mathbf{D}_{[m,n-1]},
\mathbf{D}_{[m-1,n-1]}
\right),
\end{aligned}
\end{equation}
where $\left\lVert \cdot \right\rVert_{2}$ represents the $\ell_{2}$-norm. The recurrence is subject to the standard boundary conditions, with the dynamic programming matrix initialized as $\mathbf{D}_{\left [ 1,1 \right ] } = \left \| \mathbf{X}^{(i)}_{\left [ :,1 \right ] }- \mathbf{X}^{(j)}_{\left [ :,1 \right ] }\right \|^{2}_{2}$. Finally, the DTW distance between $\mathbf{X}^{(i)}$ and $\mathbf{X}^{(j)}$ is defined as the bottom-right element of $\mathbf{D}$ as:
\begin{equation}
\label{eq2}
d_{\text{DTW}}(\mathbf{X}^{(i)}, \mathbf{X}^{(j)}) = \mathbf{D}_{[T^{(i)}, T^{(j)}]}.
\end{equation}
\subsection{Generation of Granular Balls} 
\label{subsec: Generation of Granular Balls}
GBC represents data using adaptive GBs (hyperspherical regions) rather than individual points, thereby improving the computational efficiency and robustness of downstream tasks. A GB covers several samples $\mathcal{S}^{*}$, where $\mathcal{S}^{*}\subseteq \mathcal{S}$. A GB is characterized by the center $\mathbf{C}$, the radius $r$, and the purity $p$, which are represented as follows:
\begin{align}
\mathbf{C}
&=
\operatorname*{arg\,min}_{\mathbf{X}^{(i)}\in\mathcal{S}^{*}}
\sum_{\mathbf{X}^{(j)}\in\mathcal{S}^{*}}
d_{\text{DTW}}(\mathbf{X}^{(i)},\mathbf{X}^{(j)}),
\label{eq:center}\\
r
&=
\frac{1}{|\mathcal{S}^{*}|}
\sum_{\mathbf{X}^{(i)}\in\mathcal{S}^{*}}
d_{\text{DTW}}(\mathbf{X}^{(i)},\mathbf{C}),
\label{eq:radius}\\
p
&=
\frac{
\displaystyle
\max_{y}
\left|
\left\{
i:
\mathbf{X}^{(i)}\in\mathcal{S}^{*},
\ y^{(i)}=y
\right\}
\right|
}{
|\mathcal{S}^{*}|
},\label{eq:purity}
\end{align}
where $|\cdot|$ denotes the cardinality of a set. Note that the label of a GB
is determined by the majority of samples in $\mathcal{S}^{*}$.
\par
The GBs are generated recursively in a coarse-to-fine manner. At the initial stage, all training samples are grouped into a single GB, whose center is initialized as the medoid, i.e., the sample with the minimum average DTW distance to all other samples. Each GB is then examined to determine whether it should be further partitioned. Let $GB_{+}$ denote a parent GB under examination, and let $\mathcal{S}_{+}$ and $p_{+}$ denote its sample set and purity, respectively. If the purity of $GB_{+}$ is lower than the predefined threshold and it contains a sufficient number of samples, $GB_{+}$ is split into $\kappa$ child GBs as follows:
\begin{align}
\label{eq4}
GB_{+}
\overset{\mathrm{split}}{\longrightarrow}
\left\{
GB_{-}^{(1)},\ldots,GB_{-}^{(\kappa)}
\right\}
\quad
\text{if }
p_{+}<\rho
\land
\left|\mathcal{S}_{+}\right|>\beta,
\end{align}
where $GB_{-}^{(1)},\ldots,GB_{-}^{(\kappa)}$ denote the resulting $\kappa$ child granular balls, and $\rho$ and $\beta$ are the purity threshold and the minimum sample-size threshold for splitting, respectively. In this work, we provide two strategies for selecting centers in the splitting procedure as follows:
\begin{itemize}
    \item The random strategy: $\kappa$ samples are randomly selected in $GB_{+}$ as centers.
    \item The label-informed strategy: the center of the parent granular ball is first retained as the center of $GB_{-}^{(1)}$. Subsequently, one sample is randomly selected from each class other than the class of the parent center, yielding the remaining ($\kappa-1$) child-ball centers.
\end{itemize}
Eventually, all GBs that do not satisfy the splitting criterion defined in Equation~(\ref{eq4}) are retained to form the final granular-ball set.
\subsection{Granular Ball based Nearest-Neighbor Classification} 
\label{sub:knn}
The $K$-nearest neighbors ($K$-NN) algorithm is a non-parametric supervised learning method that classifies a new sample according to the majority label among its $K$ nearest training samples~\cite{cover1967nearest}. As a lazy learning method, $K$-NN postpones most computations until the inference stage and therefore does not require an explicit model training process. In DTW-GBC, we employ $1$-NN classification ($K=1$) at the granular-ball level. Given a set of $G$ granular balls, denoted by $\mathcal{G}=\left\{GB^{(g)}\right\}_{g=1}^{G}$, the distance between the $i$-th test sample and the $g$-th granular ball is defined as
\begin{equation}
    \operatorname{dist}\left(\mathbf{X}^{(i)},GB^{(g)}\right)
    =
    d_{\mathrm{DTW}}\left(\mathbf{X}^{(i)},\mathbf{C}^{(g)}\right)
    -
    r^{(g)},
\end{equation}
where $\mathbf{C}^{(g)}$ and $r^{(g)}$ denote the center and radius of the $g$-th granular ball, respectively. For each test sample, its distance to every granular ball is calculated, and the ball with the minimum distance is selected for prediction. The majority label of the selected granular ball is then assigned to the test sample as its predicted label.
\section{Computational complexity analysis}
\label{sec: complexity analysis}
Let $N_{tr}$ and $N_{te}$ denote the numbers of training and test samples, respectively. For simplicity, we assume that all samples have the same sequence length, denoted by $T$. The detailed procedure for constructing granular balls of our proposed method is presented in Algorithm~\ref{alg:dtw_gb}. The training complexity of the proposed DTW-GBC consists mainly of two components: pairwise DTW distance computation among the $N_{tr}$ training samples and recursive granular-ball splitting. The former costs $\mathcal{O}\!\left(N_{tr}^{2}T^{2}\right)$ when standard DTW is used, corresponding to Procedures 1--5 in Algorithm~\ref{alg:dtw_gb}. Let $D$ denote the maximum depth of the splitting process. The computational overhead of recursive granular-ball splitting, corresponding to Procedures 9--19, is conservatively upper-bounded by $\mathcal{O}\!\left(DN_{\mathrm{tr}}^{2}\right)$. Therefore, the overall complexity of DTW-based granular-ball generation is
\begin{equation}
\mathcal{O}\!\left(
N_{tr}^{2}T^{2}
+
DN_{tr}^{2}
\right)
=
\mathcal{O}\!\left(
N_{tr}^{2}\left(T^{2}+D\right)
\right).
\end{equation}
Under the practical condition that $D \ll T^{2}$, the pairwise DTW computation dominates the training cost, and the overall complexity can be simplified to $\mathcal{O}\!\left(N_{tr}^{2}T^{2}\right)$. The computational complexity during the inference period is $\mathcal{O}\left ( N_{te}GT^{2} \right ) $. Compared with the 1-NN classifier using DTW distance, DTW-GBC requires additional computation to construct granular balls, but reduces the inference complexity by comparing each test sample only with the $G$ granular ball centers rather than $N_{tr}$ training samples.
\begin{algorithm}[htbp]
\caption{DTW-based Granular Ball Generation}
\label{alg:dtw_gb}
\begin{algorithmic}[1]

\renewcommand{\algorithmicrequire}{\textbf{Input:}}
\renewcommand{\algorithmicensure}{\textbf{Output:}}

\REQUIRE training dataset
$\mathcal{S}_{\mathrm{tr}}
=
\{(\mathbf{X}_{\mathrm{tr}}^{(i)},y_{\mathrm{tr}}^{(i)})\}_{i=1}^{N_{\mathrm{tr}}}$,
purity threshold $\rho$, and size threshold $\beta$.

\ENSURE a set of GBs
$\mathcal{G}=\{GB^{(g)}\}_{g=1}^{G}$.

\FOR{$i=1$ to $S_{tr}$}
\FOR{$j=i+1$ to $S_{tr}$}
\STATE Compute the DTW distance matrix $\mathbf{D}$ between $\mathbf{X}^{(i)}$ and $\mathbf{X}^{(j)}$.
\ENDFOR
\ENDFOR
\STATE Treat all training samples as an initial GB.
\STATE Initialize $\mathcal{Q}$ with the initial GB.
\STATE Initialize $\mathcal{G}\leftarrow\emptyset$.

\WHILE{$\mathcal{Q}$ is not empty}

    \STATE Remove a GB from $\mathcal{Q}$ and denote it by $GB_{+}$.
    \STATE Compute its center, radius, label, and purity $p_{+}$.

    \IF{$p_{+}\geq\rho$ \OR $|GB_{+}|\leq\beta$}
        \STATE Add $GB_{+}$ to $\mathcal{G}$.
    \ELSE
        \STATE Determine $\kappa$ and initialize the centers according to the selected splitting strategy.
        \STATE Assign each sample to its nearest center.
        \STATE Generate $\kappa$ child GBs $GB_{-}^{(1)},\ldots,GB_{-}^{(\kappa)}$ and add them to $\mathcal{Q}$.
        
    \ENDIF

\ENDWHILE

\STATE \textbf{return} $\mathcal{G}$.

\end{algorithmic}
\end{algorithm}
\section{Experiments}
\label{Sec: Experiments}
\subsection{Experimental datasets and task settings}
\label{Sec: experimental datasets}
We conducted experiments on four benchmark time-series classification datasets, including two univariate datasets selected from the UCR2018 archive~\cite{dau2019ucr} and two multivariate datasets adopted from the Multiverse archive~\cite{middlehurst2026multiverse}. These datasets cover different numbers of classes, sample sizes, sequence lengths, and data dimensions. Detailed statistics of the benchmark datasets are summarized in Table~\ref{tab:datasets}.
\par
To evaluate the robustness of the proposed method against noisy labels, we introduced symmetric label noise~\cite{northcutt2021confident} into the training set while keeping the test set unchanged. Let $L$ denote the number of classes and $\eta$ denote the predefined noise rate. For a training sample with the true label $y=i$, its noise-contaminated label $\tilde{y}$ is generated as follows:
\begin{equation}
P(\tilde{y}=j \mid y=i)=
\begin{cases}
1-\eta, & \text{if } j=i, \\[2mm]
\dfrac{\eta}{L-1}, & \text{if } j\neq i.
\end{cases}
\label{eq:symmetric_label_noise}
\end{equation}
Accordingly, each label remains unchanged with probability $1-\eta$, whereas a corrupted label is uniformly reassigned to one of the other ($L-1$) classes. To account for variability introduced by stochastic label corruption, we generate ten independently contaminated versions of the training set for each dataset. The mean values and standard deviations of the evaluation metrics on each test set are reported in the Section~\ref{Sec: experimental results}.
\begin{table*}[htbp]
\centering
\caption{Overview of experimental datasets}
\label{tab:datasets}
\begin{tabular}{|l|l|l|l|l|l|l|}
\hline
                          & \#Train & \#Test & \#Classes & \#Timesteps & \#Dimension & Source     \\ \hline
SyntheticControl          & 300     & 300    & 6         & 60          & 1           & UCR2018    \\
ECG5000                   & 500     & 4500   & 5         & 140         & 1           & UCR2018    \\
JapaneseVowels            & 270     & 370    & 9         & 7-29        & 12          & Multiverse \\
ArticularyWordRecognition & 275     & 300    & 25        & 144         & 9           & Multiverse \\ \hline
\end{tabular}
\end{table*}
\subsection{Model and parameter settings}
\label{Sec: experimental settings}
We denote DTW-GBC using the random splitting strategy and that using the label-informed splitting strategy by DTW-GBC-R and DTW-GBC-L, respectively. To investigate whether these two variants achieve greater robustness to noisy labels than their original DTW-based counterpart in terms of classification performance, we employed the conventional DTW classifier as the baseline. Regarding the hyperparameter settings, we fixed $\beta$ to 1 for two DTW-GBC based methods and searched $\rho$ in $\left \{ 0.5, 0.6, 0.7, 0.8, 0.9, 1 \right \}$. For DTW-GBC-R, the candidate values of $\kappa$ for each dataset are listed in Table~\ref{tab:searching candidates}, and the optimal value was selected using five-fold cross-validation on the training set.
\begin{table}[htbp]
\centering
\caption{Candidate values of $\kappa$ on the DTW-GBC-R for each dataset.}
\label{tab:searching candidates}
\begin{tabular}{|l|l|}
\hline
                          & Searching candidates                  \\ \hline
                          \hline
SyntheticControl          & $\left \{ 2, 4, 6 \right \}$          \\
ECG5000                   & $\left \{ 2, 3, 4, 5 \right \}$       \\
JapaneseVowels            & $\left \{ 2, 3, 6, 9\right \}$           \\
ArticularyWordRecognition & $\left \{ 2, 5, 10, 15, 20, 25\right \}$ \\ \hline
\end{tabular}
\end{table}
\subsection{Metrics}
\label{Sec: experimental metrics}
We employed accuracy, the weighted F1-score, and the weighted geometric mean (G-mean) to evaluate the classification performance. These metrics are formulated as follows:
\begin{equation}
\mathrm{Accuracy}=
\frac{1}{S_{te}}
\sum_{i=1}^{S_{te}}
\mathbb{I}\left(y_i=\hat{y}_i\right),
\label{eq:accuracy}
\end{equation}

\begin{equation}
\mathrm{F1}_{\mathrm{wtd}}=
\sum_{l=1}^{L}
w_l
\frac{2P_lR_l}{P_l+R_l},
\label{eq:weighted_f1}
\end{equation}
\begin{equation}
\mathrm{G\text{-}mean}_{\mathrm{wtd}}
=
\sqrt{
\sum_{l=1}^{L}w_l\mathrm{Sen}_l
\sum_{l=1}^{L}w_l\mathrm{Spe}_l
}.
\label{eq:weighted_gmean}
\end{equation}
where $\hat{y}^{(i)}$ is the predicted label of the $i$-th test sample and $\mathbb{I}(\cdot)$ is the indicator function, which equals 1 if $y^{(i)}=\hat{y}^{(i)}$ and 0 otherwise. The symbol $w_l=n_l/N_{te}$ is the proportion of samples belonging to class $l$. Moreover, $P_l$, $R_l$, $\mathrm{Sen}_l$, and $\mathrm{Spe}_l$ denote the precision, recall, sensitivity, and specificity of class $l$, respectively.
\begin{table*}[htbp]
\caption{Classification performances on four datasets with $\eta=0.1$ and $\eta=0.2$.}
\centering
\label{tab:classification performances_noisy}
\begin{tabular}{|l|l|c|c|c|c|c|c|} 
\hline
                                           &                                       & \multicolumn{2}{c|}{DTW}                                    & \multicolumn{2}{c|}{DTW-GBC-R}                                      & \multicolumn{2}{c|}{DTW-GBC-L}                                       \\ 
\hline
Datasets                                   & Metrics                               & $\eta=0.1$ & {\cellcolor[rgb]{0.922,0.922,0.922}}$\eta=0.2$ & $\eta=0.1$     & {\cellcolor[rgb]{0.922,0.922,0.922}}$\eta=0.2$     & $\eta=0.1$     & {\cellcolor[rgb]{0.922,0.922,0.922}}$\eta=0.2$      \\ 
\hline
\multirow{3}{*}{SyntheticControl}          & Accuracy                              & 0.900      & {\cellcolor[rgb]{0.922,0.922,0.922}}0.796      & 0.958          & {\cellcolor[rgb]{0.922,0.922,0.922}}0.948          & \textbf{0.962} & {\cellcolor[rgb]{0.922,0.922,0.922}}\textbf{0.960}  \\
                                           & $\mathrm{F1}_\mathrm{wtd}$            & 0.900      & {\cellcolor[rgb]{0.922,0.922,0.922}}0.796      & 0.958          & {\cellcolor[rgb]{0.922,0.922,0.922}}0.947          & \textbf{0.962} & {\cellcolor[rgb]{0.922,0.922,0.922}}\textbf{0.960}  \\
                                           & $\mathrm{G\text{-}mean}_\mathrm{wtd}$ & 0.939      & {\cellcolor[rgb]{0.922,0.922,0.922}}0.874      & 0.974          & {\cellcolor[rgb]{0.922,0.922,0.922}}0.968          & \textbf{0.977} & {\cellcolor[rgb]{0.922,0.922,0.922}}\textbf{0.976}  \\ 
\hline
\multirow{3}{*}{ECG5000}                   & Accuracy                              & 0.835      & {\cellcolor[rgb]{0.922,0.922,0.922}}0.742      & \textbf{0.913} & {\cellcolor[rgb]{0.922,0.922,0.922}}\textbf{0.902} & 0.907          & {\cellcolor[rgb]{0.922,0.922,0.922}}0.894           \\
                                           & $\mathrm{F1}_\mathrm{wtd}$            & 0.856      & {\cellcolor[rgb]{0.922,0.922,0.922}}0.786      & \textbf{0.908} & {\cellcolor[rgb]{0.922,0.922,0.922}}\textbf{0.884} & 0.904          & {\cellcolor[rgb]{0.922,0.922,0.922}}0.884           \\
                                           & $\mathrm{G\text{-}mean}_\mathrm{wtd}$ & 0.887      & {\cellcolor[rgb]{0.922,0.922,0.922}}0.827      & \textbf{0.929} & {\cellcolor[rgb]{0.922,0.922,0.922}}\textbf{0.911} & 0.927          & {\cellcolor[rgb]{0.922,0.922,0.922}}0.909           \\ 
\hline
\multirow{3}{*}{JapaneseVowels}            & Accuracy                              & 0.848      & {\cellcolor[rgb]{0.922,0.922,0.922}}0.759      & \textbf{0.898} & {\cellcolor[rgb]{0.922,0.922,0.922}}0.871          & 0.897          & {\cellcolor[rgb]{0.922,0.922,0.922}}\textbf{0.881}  \\
                                           & $\mathrm{F1}_\mathrm{wtd}$            & 0.849      & {\cellcolor[rgb]{0.922,0.922,0.922}}0.760      & \textbf{0.899} & {\cellcolor[rgb]{0.922,0.922,0.922}}0.869          & 0.898          & {\cellcolor[rgb]{0.922,0.922,0.922}}\textbf{0.880}  \\
                                           & $\mathrm{G\text{-}mean}_\mathrm{wtd}$ & 0.912      & {\cellcolor[rgb]{0.922,0.922,0.922}}0.857      & \textbf{0.941} & {\cellcolor[rgb]{0.922,0.922,0.922}}0.924          & 0.940          & {\cellcolor[rgb]{0.922,0.922,0.922}}\textbf{0.930}  \\ 
\hline
\multirow{3}{*}{ArticularyWordRecognition} & Accuracy                              & 0.895      & {\cellcolor[rgb]{0.922,0.922,0.922}}0.789      & 0.938          & {\cellcolor[rgb]{0.922,0.922,0.922}}0.883          & \textbf{0.955} & {\cellcolor[rgb]{0.922,0.922,0.922}}\textbf{0.929}  \\
                                           & $\mathrm{F1}_\mathrm{wtd}$            & 0.894      & {\cellcolor[rgb]{0.922,0.922,0.922}}0.786      & 0.936          & {\cellcolor[rgb]{0.922,0.922,0.922}}0.876          & \textbf{0.953} & {\cellcolor[rgb]{0.922,0.922,0.922}}\textbf{0.923}  \\
                                           & $\mathrm{G\text{-}mean}_\mathrm{wtd}$ & 0.944      & {\cellcolor[rgb]{0.922,0.922,0.922}}0.884      & 0.967          & {\cellcolor[rgb]{0.922,0.922,0.922}}0.937          & \textbf{0.976} & {\cellcolor[rgb]{0.922,0.922,0.922}}\textbf{0.962}  \\
\hline
\end{tabular}
\end{table*}
\subsection{Results}
\label{Sec: experimental results}
The classification results of the three methods at $\eta=0.1$ and $\eta=0.2$ are reported in Table~\ref{tab:classification performances_noisy}, respectively. Note that $\kappa=2$ was selected for DTW-GBC-R, as it consistently achieved the best validation performance across all datasets. At both noise levels, DTW-GBC-R and DTW-GBC-L outperform conventional DTW on all datasets and evaluation metrics. When $\eta=0.1$, DTW-GBC-L performs best on SyntheticControl and ArticularyWordRecognition, while DTW-GBC-R performs slightly better on ECG5000 and JapaneseVowels. When $\eta=0.2$, DTW-GBC-L achieves the best results on three of the four datasets, with DTW-GBC-R performing best on ECG5000. The improvement over conventional DTW is particularly evident at the higher noise rate. These results indicate that replacing individual training samples with granular-ball-level representations reduces the sensitivity of DTW-based 1-NN classification to corrupted training labels, thereby improving robustness under label noise.
\par
\begin{figure*}[htbp]
    \centering
    \includegraphics[scale=0.27]{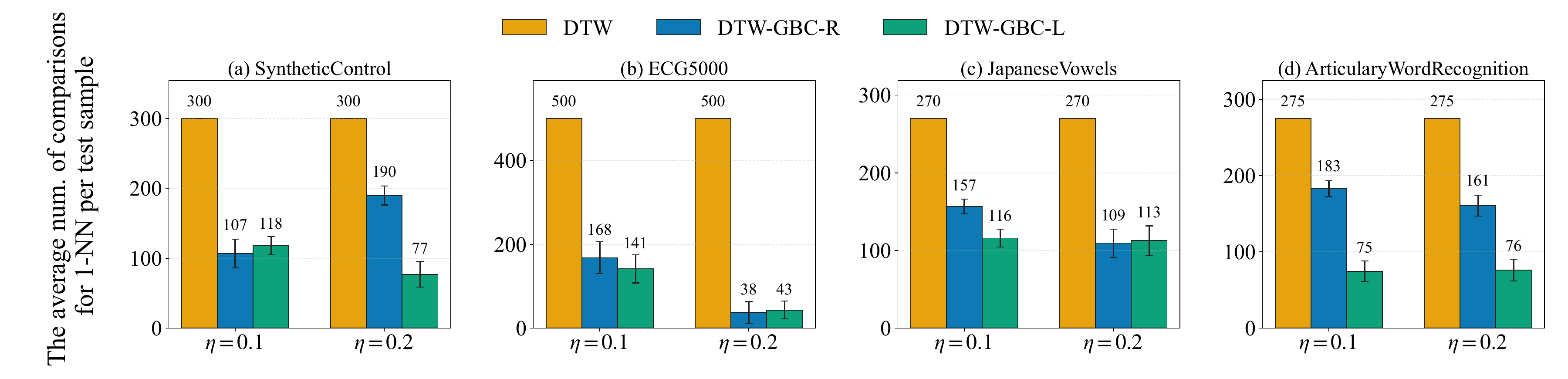}
    \caption{The average numbers of 1-NN comparisons per test sample corresponding to the best performances for each method reported in Table~\ref{tab:classification performances_noisy}.}
    \label{fig:2}
\end{figure*}

Figure~\ref{fig:2} shows the average numbers of comparisons corresponding to the best-performing settings of the three methods reported in Table~\ref{tab:classification performances_noisy}. Since conventional DTW performs 1-NN classification by comparing each test sample with every training sample, it requires 300, 500, 270, and 275 comparisons per test sample on SyntheticControl, ECG5000, JapaneseVowels, and ArticularyWordRecognition, respectively. In contrast, the two DTW-GBC variants perform 1-NN classification over the generated granular balls, and thus the number of comparisons is directly determined by the number of generated balls. As shown in Figure~\ref{fig:2}, DTW-GBC-R and DTW-GBC-L require an average of only 38--190 comparisons per test sample across the four datasets and two noise levels, corresponding to reductions of approximately 33.5\%--92.4\% compared with conventional DTW. Overall, DTW-GBC improves robustness to label noise while substantially reducing the number of DTW comparisons required during inference.
\par
\begin{figure*}[htbp]
    \centering
    \includegraphics[scale=0.25]{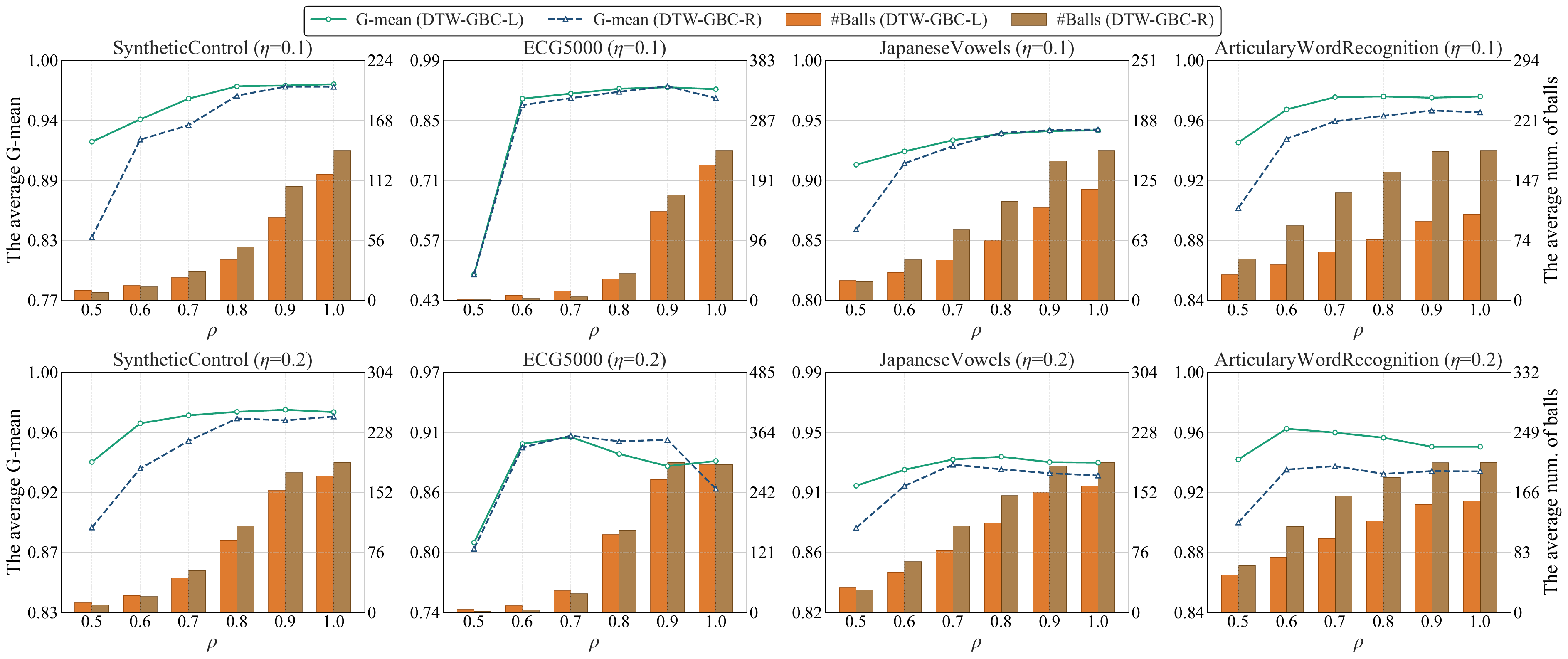}
    \caption{The average weighted G-mean and average number of generated granular balls of DTW-GBC-R and DTW-GBC-L under different $\rho$ on four benchmark datasets with $\eta=0.1$ (upper) and $\eta=0.2$ (bottom). }
    \label{fig:3}
\end{figure*}

To further investigate how the granular-ball construction affects both classification performance and inference efficiency, Figure~\ref{fig:3} shows the average weighted G-mean and the average number of generated balls for DTW-GBC-R and DTW-GBC-L under different $\rho$. As $\rho$ increases, both variants generally generate more granular balls because a stricter purity requirement causes more parent balls to be recursively split. In contrast, the weighted G-mean typically improves rapidly at relatively small values of $\rho$ and then tends to saturate, or even decrease slightly on some datasets. This indicates that generating increasingly fine-grained granular balls does not necessarily yield further classification improvement, while it directly increases the number of comparisons required during inference.
\par
Moreover, under the same value of $\rho$, DTW-GBC-L generally generates fewer granular balls than DTW-GBC-R while achieving comparable or higher weighted G-mean. This tendency is especially evident on JapaneseVowels and ArticularyWordRecognition, where the label-informed strategy maintains higher classification performance with a more compact granular representation over a wide range of $\rho$. The difference is less consistent on ECG5000 when $\eta=0.2$, indicating that the advantage of label-informed splitting can depend on the characteristics and noise level of the dataset. These results show that DTW-GBC provides a favorable trade-off between robustness to noisy labels and inference efficiency, while the label-informed splitting strategy generally produces a more compact granular representation.

\section{Conclusion}
\label{Sec: Discussion}
This study proposed a DTW-based granular-ball computing framework for robust and efficient time-series classification under noisy labels. By organizing temporally similar training samples into granular balls and performing classification at the granule level, DTW-GBC helps reduce the sensitivity of nearest-neighbor classification to individual mislabeled samples. Experiments on four benchmark datasets demonstrate that both DTW-GBC variants outperform conventional DTW under symmetric label noise in most cases. Among the two variants, DTW-GBC-L generally achieves better classification performance and provides a favorable performance--efficiency trade-off through label-informed splitting. Moreover, both variants substantially reduce the number of DTW comparisons during inference by comparing test samples with granular balls rather than all training instances. Overall, the results demonstrate that integrating GBC with DTW can improve robustness to noisy labels while reducing the computational cost of 1-NN inference.
\par
A limitation of this study is that the value of $\kappa$ for DTW-GBC-R is selected using conventional cross-validation on noisy training data. Consequently, the selected value may be sensitive to a particular realization of label noise, especially for datasets with limited training samples.
\par
In future work, in addition to exploring more reliable strategies for hyperparameter selection under noisy labels, we plan to evaluate DTW-GBC on a broader range of time-series classification datasets and investigate its robustness under other types of label noise. We also intend to extend the proposed framework to other time-series analysis tasks, such as clustering and anomaly detection.

\section*{Acknowledgment}
This work was partly supported by JSPS KAKENHI Grant Number JP25K24744, JST Moonshot R\&D Grant Number JPMJMS2021, and JST CREST Grant Number JPMJCR24R2.

\bibliographystyle{IEEEtran}
\bibliography{references.bib}
\end{document}